\documentclass{article}
\usepackage{spconf,amsmath,graphicx,hyperref}
\ninept

\usepackage{multirow}
\usepackage{bbding}
\usepackage{makecell}
\usepackage{graphicx}
\usepackage[table]{xcolor}
\usepackage{lipsum}
\usepackage{CJKutf8}
\usepackage{algorithm}
\usepackage{algorithmic}
\usepackage{url}
\usepackage{booktabs}

\title{PROGRESSIVE REFINEMENT: AN ITERATIVE PSEUDO-LABELING APPROACH FOR MANDARIN-ENGLISH CODE-SWITCHING ASR}

\name{Qu Yang$^{1,2,*}$ \thanks{$^*$ Work done during an internship at Apple. quyang@u.nus.edu}, Cakra Wardhana$^{1}$, Tim Ng$^{1}$}
\address{$^{1}$ Apple, Singapore \quad
         $^{2}$ National University of Singapore, Singapore}

\begin{document}
\begin{CJK*}{UTF8}{gbsn}

\maketitle

\begin{abstract}
Code-switching (CS), alternating languages within the same utterance, poses significant challenges for automatic speech recognition (ASR) due to limited CS training data. This paper applies an iterative pseudo-labeling training approach to CS-ASR for the first time, demonstrating its effectiveness in leveraging unlabeled data to improve CS-ASR performance.
The approach comprises three phases: pseudo-label generation, two-stage bilingual model training, and iterative improvements.
It begins by generating pseudo-labels from a large unlabeled corpus, creating a semi-supervised dataset. This dataset supports a two-stage training framework where the model is pre-trained and then fine-tuned on supervised CS data. Iterative refinements further enhance the model’s accuracy in handling complex CS scenarios.
Our approach significantly advances CS-ASR systems, achieving notable Mix Error Rate (MER) reductions
on SEAME's \textit{devman} (6.35\%) and \textit{devsge} (8.29\%) subsets.
\end{abstract}


\renewcommand{\thefootnote}{}

\renewcommand{\thefootnote}{\fnsymbol{footnote}}

\section{Introduction}

Code-switching (CS) is common in multilingual societies like Southeast Asia \cite{lyu2010seame}.
This linguistic phenomenon poses unique challenges for automatic speech recognition (ASR) systems \cite{liu2023reducing,diwan2021multilingual,modipa2013implications}, as it introduces complex variations and unpredictable switches between languages that conventional ASR models struggle to handle. Improving ASR for code-switching speech is therefore crucial to enhancing user experience in multilingual environments, particularly for voice assistant technologies \cite{li2022language}.

Code-switching ASR has gained attention recently, with researchers exploring various approaches to address its complexities. Studies have focused on language-specific architectures, such as joint CTC-attention with language identification (LID) \cite{luo2018towards, zeng2018end, shan2019investigating}, and transformer-based multi-encoder-decoder frameworks \cite{zhou2020multi}. Self-supervised pre-training with multilingual data has also shown promise for improving Mandarin-English CS-ASR \cite{chen2024self, tseng2021mandarin}. The dual-encoder transformer network \cite{lu2020bi}, which employs encoders pre-trained in Mandarin and English to extract language-specific features, has inspired further developments in this framework \cite{nj2020investigation, song2022language, dalmia2021transformer}. However, the scarcity of labeled code-switching data remains a major obstacle to advancing ASR performance for mixed-language speech.

In this work, we apply an iterative pseudo-labeling training approach to English-Mandarin code-switching ASR for the first time, demonstrating its effectiveness in improving CS-ASR performance.
While pseudo-labeling has been widely explored in ASR research \cite{likhomanenko2022continuous, berrebbi2022continuous,lugosch2022pseudo}, our approach differs in key aspects. First, those existing pseudo-labeling methods focus on improving monolingual or code-mixing \cite{poplack2003pieter} ASR where lexical items from different languages appear in the same utterance but without frequent grammatical switching. In contrast, our method specifically addresses code-switching speech which involves more dynamic and spontaneous alternation between languages.
Second, our approach leverages unlabeled data that we assume to contain code-switching scenarios, whereas prior methods typically rely on monolingual or code-mixing datasets that lack such code-switching contexts.
Finally, our initial ASR model (M0) combines Mandarin and English monolingual data during training, providing it with a robust understanding of both languages from the outset. This bilingual initialization sets our method apart from previous pseudo-labeling works.

Our proposed method consists of three phases: pseudo-label generation, two-stage bilingual model training (comprising pre-training and fine-tuning), and iterative improvements. The pseudo-label generation phase uses existing ASR models to label a vast corpus of unlabeled data, thereby creating a diverse semi-supervised training set that incorporates broader linguistic variations \cite{kahn2020self, weninger2020semi}. During the two-stage training, the model is pre-trained on pseudo-labeled data to learn general patterns before being fine-tuned on a smaller supervised dataset that includes both monolingual and code-switching utterances. This iterative training cycle enhances the model's accuracy, enabling it to handle complex code-switching scenarios. The contributions of this research are threefold:
\begin{itemize}
\item We harness the potential of unlabeled  data for model pre-training, significantly increasing the training data we can use beyond labeled datasets.

\item The iterative aspect of our training methodology allows for continuous refinement of the ASR model, utilizing newly generated pseudo-labels in each cycle to enhance learning and adaptability.

\item Our approach demonstrates substantial reductions in Mix Error Rates (MERs) on the SEAME dataset \cite{lyu2010seame}, indicating significant improvements in the model's capability to handle code-switching speech.
\end{itemize}

\section{Iterative Pseudo-Labeling Training}
\begin{figure}[htp]
    \centering
    \hspace{-5mm}
    \includegraphics[scale=0.54]{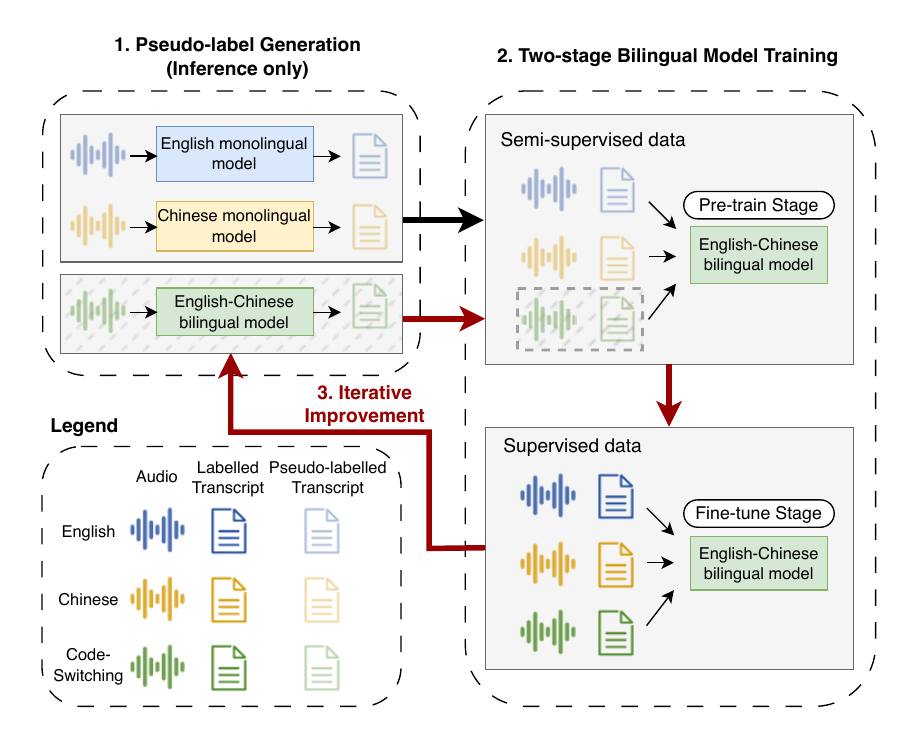}
    \caption{Overview of the iterative pseudo-labeling training approach. The pipeline consists of three key phases: (1) Pseudo-label generation,
    (2) Two-stage bilingual model training,
    and (3) Iterative improvement.
    The shaded boxes in phases (1) and (2) are only applicable after the initial bilingual model (M0) is trained and is used for subsequent iterative training (M1, M2).
    Bold arrows indicate the training process flow, while thin arrows represent data flow. Red arrows depict the iterative loop for continuous improvement.
    }
    \label{fig:overiew}
\end{figure}

This section outlines our iterative pseudo-labeling approach for improving ASR performance in code-switching environments.
The overall pipeline is depicted in Figure \ref{fig:overiew}.

\subsection{CTC+Attention Model}
The CTC+Attention model is a widely adopted architecture in ASR systems \cite{watanabe2017hybrid,kim2017joint,liu2023reducing,yan2023towards}, combining the strengths of Connectionist Temporal Classification (CTC) \cite{graves2006connectionist} and the attention mechanism \cite{vaswani2017attention,bahdanau2016end}. CTC is effective for aligning input speech frames to output tokens by modeling sequence probabilities, while the attention mechanism excels at capturing long-range dependencies and generating contextually appropriate outputs.
This architecture forms the backbone of our iterative pseudo-labeling framework, enabling precise and efficient transcription of multilingual and code-switching speech. Unlike many ASR systems \cite{liu2023reducing, liu2024aligning}, our approach does not rely on an external language model, ensuring end-to-end learning without additional decoding constraints.

\subsection{Pseudo-label Generation}
The first phase involves generating pseudo-labels from unlabeled data. We use three audio-only data: (1) monolingual English; (2) monolingual Mandarin; and (3) code-switching English-Mandarin audio from Singapore, assuming that the language environment in Singapore naturally contains code-switching interactions \cite{lyu2010seame}. Those data are identified using metadata on language identifiers embedded in the original interaction logs.

As shown in Figure \ref{fig:overiew}, we use three ASR models to generate pseudo-labels: (1) a monolingual English ASR model to pseudo-label English audio, (2) a monolingual Mandarin ASR model to pseudo-label Mandarin audio, and (3) a code-switching English-Mandarin ASR model to pseudo-label code-switching audio. An initial ASR model (M0), trained on monolingual English and Mandarin data, is used to do initial pseudo-labeling on code-switching data.

The resulting pseudo-labeled dataset forms a foundation for subsequent iterations, improving accuracy for code-switching speech. Prior research \cite{xu2020iterative} has shown that pseudo-labeling enhances monolingual ASR by increasing robustness and linguistic coverage. We hypothesize that these benefits extend to code-switching by having representative pseudo-labeled data.

\subsection{Two-stage Bilingual Model Training}
The second phase of the proposed method follows a two-stage training, which consists of pre-training and fine-tuning, as depicted in Figure \ref{alg:pseudo_label}.
Pre-training introduces diverse linguistic patterns, providing a foundational understanding of the model. The subsequent stage of fine-tuning with high-quality annotated data further improves the model’s ability to process code-switching speech.

\textbf{Pre-training Stage:} In this stage, the proposed model is pre-trained on the pseudo-labeled dataset generated from the first phase.
As depicted in Figure \ref{fig:overiew}, the model is pre-trained using pseudo-labeled transcriptions, enabling it to learn from diverse linguistic patterns present in the data.

\textbf{Fine-tuning Stage:} The pre-trained model is then fine-tuned using a smaller, human-annotated dataset that contains both monolingual and code-switching audio-text data. By focusing on high-quality labeled data, this stage helps the model converge faster and better capture the syntactic and semantic transitions in code-switching speech.
Figure \ref{fig:overiew} (bottom right) illustrates how the supervised dataset plays a critical role in fine-tuning.
The high-quality transcriptions in this stage provide stronger supervised learning guidance, which is expected to improve model performance.

\subsection{Iterative Improvement}
The final phase involves iterative refinement on pseudo-labels, where the model trained in the previous iteration generates improved pseudo-labels \cite{xu2020iterative} for code-switching audio-only data.

As shown in Figure \ref{fig:overiew} (red arrow connecting the bottom-right box to the top-left box), the iterative refinement process begins with our proposed model from the previous iteration. This model generates new pseudo-labels for the audio-only code-switching dataset, incorporating updated predictions into the training pipeline for subsequent iterations. By repeating this cycle, the model progressively refines the pseudo-labels for the audio-only code-switching dataset.

This iterative approach ensures that the model benefits from continuous refinement, gradually improving its performance with each cycle. The process leverages both the breadth of pseudo-labeled data and the precision of supervised data, making it highly effective for addressing the challenges of code-switching ASR as shown in the experiment results. The iterative training process is summarized in Algorithm \ref{alg:pseudo_label}.

\begin{algorithm}[h]
\caption{Iterative Pseudo-labeling Algorithm}
\label{alg:pseudo_label}
\begin{algorithmic}[1]
\REQUIRE $S = \{(x_i, y_i)\}$: supervised data, $U = \{x_j\}$: unlabeled data, $M_0$: initial ASR model trained on monolingual data
\STATE Initialize model $M \gets M_0$
\WHILE{the desired number of rounds for convergence has not been reached}
    \STATE Generate the pseudo-labelled set: $P = \{(x_j, M(x_j)) \mid x_j \in U\}$
    \STATE Obtain $M_1$ by:
    \STATE \hspace{1em}1) Pre-training $M$ on $P$
    \STATE \hspace{1em}2) Fine-tuning $M$ on $S$
    \STATE Replace $M$ with $M_1$
\ENDWHILE
\RETURN $M$
\end{algorithmic}
\end{algorithm}

\section{Experiments and Results}
\subsection{Datasets}
We use both monolingual and code-switching datasets, combining supervised and semi-supervised data to train the proposed system. We utilize two publicly available datasets, SEAME \cite{lyu2010seame, LDC2015S04} and NSC \cite{koh2019building}, along with private supervised datasets.

The SEAME dataset, a code-switching corpus of English and Mandarin speech collected in Malaysia and Singapore, is widely used for training and evaluating ASR systems in multilingual environments. It includes 96.6 hours of training data, 4.9 hours of validation data, and two evaluation subsets: \textit{devman} (Mandarin-dominant) and \textit{devsge} (English-dominant), providing diverse code-switching scenarios.

The enSGeval dataset is a private corpus of 15 hours of human-transcribed monolingual Singaporean English speech. This is to ensure robust performance in monolingual English scenarios. Together, these datasets enable comprehensive evaluation across code-switching and monolingual settings.

For training, we employ large-scale supervised and semi-supervised datasets. Our private monolingual semi-supervised data contains 100k hours of pseudo-labeled English (enW) and 44k hours of monolingual Mandarin (zhW) speech, while code-switching data includes 22.4k hours of pseudo-labeled English-Mandarin code-switching speech. Supervised datasets comprise NSC (part 1, 2, and 4), SEAME, and private datasets, with the final size of 11.6k hours of English (enW), 6.9k hours of Singaporean English (enSG), 13.7k hours of Mandarin (zhW), and 1k hours of English-Mandarin code-switching speech from NSC part 4 and SEAME.

\subsection{Experimental Setup}
The experimental setup for training and evaluating the proposed system involves multiple configurations to ensure a comprehensive assessment of its capabilities in handling code-switching speech.
The baseline model utilizes a CTC+Attention ASR architecture, consisting of twelve conformer encoder layers coupled with six transformer decoder layers \cite{liu2023reducing,gulati2020conformer,liu2024aligning}. It is trained on the SEAME dataset using a private framework/tooling, and it follows standard practices for code-switching ASR baselines as demonstrated in related works \cite{liu2023reducing, liu2024aligning}.

The initial model, referred to as M0, differs from the baseline in that it is a bilingual ASR model trained on pseudo-labeled monolingual English and monolingual Mandarin semi-supervised data during the pre-training phase. It is then fine-tuned on the smaller, human-annotated monolingual (English and Mandarin) datasets. Unlike the baseline, M0’s training approach combines both semi-supervised and supervised data. Notably, the training process for M0 does not involve any code-switching data.

The iterative models, M1 and M2, are developed through successive iterations of training, where each iteration leverages the updated pseudo-labels generated by the model from the previous iteration (i.e., M0 for M1 and M1 for M2, etc.). This iterative training process aims to progressively enhance the model's performance by refining the pseudo-labels, especially in recognizing code-switching utterances. The iterative nature of training allows the model to continuously improve its handling of code-switching English-Mandarin speech.

The pre-training phase employs a batch size of $48$ and involved $150$k training steps, with a warm-up period of $20$k steps. The fine-tuning phase uses a batch size of $64$, with $100$k training steps and a warm-up period of $10$k steps. The model vocabulary consists of $14$k tokens, combining English subword tokens and Chinese characters as single-character SentencePiece tokens (both using a unigram approach \cite{kudo2018subword}).

\subsection{Results and Analysis}
\subsubsection{Overall Performance}

\begin{table}[t]
\centering
\caption{Performance comparison among the baseline model, monolingual (private) model, and models trained using the proposed iterative pseudo-labeling approach.}
\label{tab:overall}
\begin{tabular}{lccccccc}
\toprule
\multicolumn{1}{l}{\multirow{2}{*}{\textbf{Model}}} & \multicolumn{2}{c}{SEAME (MER)} & {enSGeval} \\
\multicolumn{1}{c}{}                                & \textit{devman}           & \textit{devsge}           & (WER)                               \\
\midrule
\textbf{Baseline}                & 19.23            & 27.18            & 83.54                        \\
\textbf{Private model}                                     & 85.31            & 64.53            & 13.80                             \\
\textbf{Initial bilingual, M0}                      & 61.09            & 54.12            & 13.22                          \\
\textbf{First Iterative, M1}                        & 13.39            & 19.47            & 12.86                          \\
\textbf{Subsequent Iterative, M2}                   & 12.88            & 18.89            & 12.89                          \\
\bottomrule
\end{tabular}
\end{table}

\begin{table*}[h]
\centering
\caption{Decoding examples. The utterance is drawn from the SEAME, where error tokens are in red.}
\label{tab:decoding}
\resizebox{1.0 \linewidth}{!}{
\begin{tabular}{lll}
\toprule
                                                    & \textbf{\textit{devman}}        & \textbf{\textit{devsge}}        \\
\midrule
\textbf{Ground Truth}       & ref: malaysia 不\ 够\ 钱\ pump 进\ 去\ 一\ 个\ fully research 的\ private company or 什\ 么\             & ref: 在\ 那\ 个\ 房\ 间\ 里\ 面\ really everything was as if like             \\
\midrule
\textbf{Private model}             & hyp: \textcolor{red}{measure booko Chian pumpkining sheet it could} fully research \textcolor{red}{the} private company \textcolor{red}{also}            & hyp: \textcolor{red}{technical and} really everything was as $\textcolor{red}{**}$         \\
\midrule
\textbf{M0}                 & hyp: \textcolor{red}{事}\ 不\ 够\ 钱\ \textcolor{red}{pumping chi could} fully research the private company \textcolor{red}{also}                         & hyp: 在\ 那\ 个\ 房\ 间\ 里\ 面\ really everything was as if $\textcolor{red}{*}$        \\
\midrule
\textbf{M1}                 & hyp: malaysia 不\ 够\ 钱\ pump 进\ 去\ 一\ 个\ fully research 的\ private company or 什\ 么\             & hyp: 在\ 那\ 个\ 房\ 间\ 里\ 面\ really everything was as if $\textcolor{red}{*}$             \\
\midrule
\textbf{M2}                 & hyp: malaysia 不\ 够\ 钱\ pump 进\ 去\ 一\ 个\ fully research 的\ private company or 什\ 么\             & hyp: 在\ 那\ 个\ 房\ 间\ 里\ 面\ really everything was as if like             \\
\bottomrule
\end{tabular}}
\end{table*}

Evaluation is done using Mix Error Rate (MER), which encompasses both the Word Error Rate (WER) for English and Character Error Rate (CER) for Mandarin.
As summarized in Table \ref{tab:overall}, the iterative models (M1 and M2) show significant improvements in MER on SEAME. The initial model (M0) has MERs of 61.09\% on \textit{devman} and 54.12\% on \textit{devsge}. With iterative training, M1 and M2 progressively reduce these MERs. M2 shows the best performance, reducing MER to 12.88\% on \textit{devman} and 18.89\% on \textit{devsge}, outperforming the baseline (19.23\%/27.18\%). These results highlight the effectiveness of iterative pseudo-labeling in refining code-switching performance.

The enSGeval dataset is used to assess performance on monolingual English speech from Singapore, ensuring no regression while optimizing for code-switching.
As shown in Table \ref{tab:overall}, our private monolingual model serves as a baseline with WER of 13.80\%. M0 achieves WER of 13.22\%, with M1 and M2 improving to 12.86\% and 12.89\%, respectively, confirming improvement on code-switching while maintaining strong performance on monolingual data.

Table \ref{tab:decoding} provides qualitative decoding examples that highlight the improvements achieved through iterative training. M0 produces several errors in both languages, particularly in mixed-language contexts, as seen in the incorrect transcription of terms like "measure booko" and "pumping chi." By M2, these errors are significantly reduced, and the model accurately captures challenging sequences such as "malaysia 不\ 够\ 钱\ pump 进\ 去\ 一\ 个." These examples illustrate how iterative refinement enhances the model's ability to handle complex linguistic transitions in code-switching speech.

\subsubsection{Ablation Study on Training Strategy}
We conduct an ablation study to evaluate the effectiveness of different training strategies for combining semi-supervised and supervised learning in our two-stage bilingual model framework. Table \ref{tab:ab_stratergy} summarizes the results of four configurations and their impact on MER on SEAME.

First, a single-stage strategy using only supervised data (M1.b) is tested. While isolating the effect of supervised learning, this approach struggles with code-switching with worse MER. Next, combining pseudo-labeled and supervised data in a single-stage strategy (M1.c) improves performance compared to M1.b but still falls short of the two-stage strategies, highlighting the importance of the pre-training stage.

The two-stage strategies (in M1.d and M1.e) both are pre-trained on semi-supervised monolingual data but differ in fine-tuning stage. M1.d is fine-tuned on supervised data first, followed by semi-supervised code-switching data. M1.e reverses this order, by fine-tuning it first with code-switching semi-supervised and then with supervised data.
M1.e achieves superior MERs of 13.30\% on \textit{devman} and 19.17\% on \textit{devsge}.
In contrast, M1.d suffers from higher MERs, likely due to fine-tuning on noisy pseudo-labeled data.

\begin{table}[t]
\centering
\caption{MER results for ablation study on training strategy.}
\label{tab:ab_stratergy}
\resizebox{\linewidth}{!}{
\begin{tabular}{llc}
\toprule
\textbf{Model}  & \textbf{Training Strategy}    & \textbf{\textit{devman}/\textit{devsge} } \\
\midrule
M1              & Two-stage; semi- and supervised     & 13.39   / 19.47        \\
\midrule
M1.b            & Single-stage; supervised only       & 15.31   / 21.49         \\
M1.c            & Single-stage; semi- and supervised  & 14.75   / 20.71         \\
\midrule
M1.d            & Two-stage; semi- after supervised   & 42.79   / 46.06          \\
M1.e            & Two-stage; semi- before supervised  & 13.30   / 19.17          \\
\bottomrule
\end{tabular}}
\end{table}

\subsubsection{Ablation Study on Sampling Weights}
This study examines the effect of sampling weights during fine-tuning, which control the proportion of data from each dataset in one batch during one training step. For instance, a 10:1 ratio means one dataset is 10 times more likely to get sampled than another. Table \ref{tab:ab_weighting} shows MER results for \textit{devman} and \textit{devsge} subsets under different weighting strategies.

In M1, sampling weights of 10 for enSG, 1 for enW and zhW, and 5 for SEAME yields MERs of 13.39\% and 19.47\%. Increasing SEAME's weight to 10 (M1.f) slightly raises MERs to 13.59\% and 19.71\%, while a further increase to 100 (M1.g) degrades the performance to 14.12\% and 20.69\%. These results suggest that overemphasizing code-switching data harms generalization, particularly for monolingual-dominated utterances.

In M1.h, adjusting the weight proportional to each dataset size results in significantly worse MERs (17.16\% and 23.28\%), highlighting the need to account for the unique roles of monolingual and code-switching data. Similarly, M1.i, with equal weights of 1, performs slightly worse (MERs of 13.42\% and 19.63\%) compared to the optimized weighted configuration (M1).

The findings emphasize the importance of carefully tuned sampling weights. Prioritizing code-switching data with moderate emphasis, while balancing monolingual contributions, optimizes bilingual ASR accuracy on code-switching speech.

\begin{table}[h]
\centering
\caption{MER result for ablation study on sampling weight}
\label{tab:ab_weighting}
\resizebox{0.9 \linewidth}{!}{
\begin{tabular}{llc}
\toprule
\textbf{Model}          & \textbf{Sampling Weight}       & \textbf{\textit{devman}/\textit{devsge} }   \\
\midrule
\multirow{3}{*}{M1}     & enW, zhW = 1      & \multirow{3}{*}{13.39   / 19.47}      \\
                        & enSG = 10         & \\
                        & SEAME = 5         & \\
\midrule
\multirow{3}{*}{M1.f}   & enW, zhW = 1      & \multirow{3}{*}{13.59   / 19.71}      \\
                        & enSG = 10         & \\
                        & SEAME = 10         & \\
\midrule
\multirow{3}{*}{M1.g}   & enW, zhW = 1      & \multirow{3}{*}{14.12   / 20.69}      \\
                        & enSG = 10         & \\
                        & SEAME = 100         & \\
\midrule
M1.h            & Proportional to dataset size     & 17.16   / 23.28      \\
\midrule
M1.i            & All set to 1              & 13.42   / 19.63      \\
\bottomrule
\end{tabular}}
\end{table}

\section{Conclusion}
This study introduces the application of an iterative pseudo-labeling approach for code-switching, achieving substantial MER reductions on SEAME while maintaining strong monolingual performance on enSGeval. To our best knowledge, this is the first work to apply iterative pseudo-labeling to code-switching ASR, demonstrating its effectiveness in complex mixed-language scenarios. By leveraging vast amount of unlabeled code-switching data effectively, the proposed method addresses the scarcity of code-switching data and achieves good accuracy improvements.

\bibliographystyle{IEEEtran}
\bibliography{mybib}

\end{CJK*}
\end{document}